# Analysis of hydrocyclone performance based on information granulation theory

**Hamed Owladeghaffari [1], Majid Ejtemaei\* [2] and Mehdi Irannajad [3]**

[1] Dept.mining & metallurgical engineering,Teharn,Iran
h.o.ghaffari@gmail.com

[2] Dept.mining & metallurgical engineering,Teharn,Iran
majidejtemaei@gmail.com

[3] Dept.mining & metallurgical engineering,Teharn,Iran
irannajad@aut.ac.ir

**ABSTRACT.** This paper describes application of information granulation theory, on the analysis of hydrocyclone perforamance. In this manner, using a combining of Self Organizing Map (SOM) and Neuro-Fuzzy Inference System (NFIS), crisp and fuzzy granules are obtained(briefly called SONFIS). Balancing of crisp granules and sub fuzzy granules, within non fuzzy information (initial granulation), is rendered in an open-close iteration. Using two criteria, "simplicity of rules "and "adaptive threoshold error level", stability of algorithm is guaranteed. Validation of the proposed method, on the data set of  the hydrocyclone is rendered.

## 1. Introduction

In the design of mineral processing cycle, one of the most important issues is the selection of hydrocyclone in different parts of the site.  However, prediction of hydrocyclone performance using direct or indirect modeling has an own difficulties. Apart from analytical, numerical, or experimental modeling, modeling based on intelligent systems can be supposed as an excellent situation, which is ensued by data engineering, machine learning, and stochastic learning theorems.  In recent years employment of these methods, whether in data analysis or control process in mineral engineering, has extended.  As instance, Karr et all (2000), have applied genetic algorithm to optimize the design and performance of hydrocyclones.  The used genetic algorithm pursued two objects: tune a fuzzy logic and deriving a genetic algorithm seeking to optimize of hydrocyclone circuit performance by incorporating of the fuzzy model.
The main disadvantages of such model, is that did not take in to account the superfluous and abnormal data set (because of inseparable feature of uncertainty and vagueness in pilot plant or laboratory monitoring (or testing)).
Employing of similar models of fuzzy logic (fuzzy inference system) and neural network, in cone crusher control (Moshghbar et al 1995) and control of flotation column (Carvalho et al 2002 & Viehia et al 2005) have been pointed.
With advancing and extension of intelligent knowledge discovery (Data mining), in different applied sciences, selection of best features, accounting of the vagueness and roughness of the monitored data, are the main challenges of the most sciences.
Because of being the uncertainty feature of the monitored data, accounting of uncertainty approaches such probability, fuzzy set and rough set theories to knowledge acquisition, extraction of rules and prediction of unknown cases, have been distinguished, more than the past. For example, Zadeh (2005) has pointed the role of fuzzy set theory in system theory (especially in control) will be increased during future years. The granulation of information theory (Zadeh, 1997) covers the mentioned approaches in two formats: crisp information granulation and fuzzy information granulation. There are two reasons why we propose this concept to tackle uncertainty in the monitored mineral processing data. The first one is human instinct. As human beings, we have developed a granular view of the world. When describing a problem, we tend to shy away from numbers and use aggregates to ponder the question instead. This is especially true when a problem involves incomplete, uncertain, or vague information. It may be sometimes difficult to differentiate distinct elements, and so one is force consider "information granules (IG) which one collection of entities arranged together due to their similarity, functional adjacency, and indistinguishability (Yo &Yao, 2002).

The process of constructing IGs is referred to as information granulation and emphasized the fact that a plethora of details doesn't necessity amount to knowledge. This was first pointed out in the pioneering work of Zadeh(1979).

Granulation serves as an abstraction mechanism for reducing an entire conceptual burden. By changing the size of the IGS, we can hide, or reveal more or less details (Bargiela & Pedrycz, 2003). The second reason is about the behavior of data. In many practical data sets, such as mineral processing engineering, the normal group and abnormal group are considered separate populations.

If we construct IGs by the similarity of numerical data, the amount of IGs in normal group will be remarkably smaller than the size of normal numerical data.

In other words, if we consider IGs instead of numerical data, it might increase the proportion of abnormal data (Su et al, 2006).

In this study, using two computational intelligence (CI) theories, neural networks, and fuzzy inference system, based on information granulation theory, an algorithm to analyses hydrocyclone data will be presented. In this model, self-organizing feature map, Neuro-Fuzzy Inference System is utilized to construct IGs. To determine suitable granulation level, the two criteria, "simplicity of rules "and "suitable error level", are supposed.

Reset of paper has been organized as follow: section 2 covers a brief review on construction of information granules. Next section, describes proposed method. Finally, in section 4, the actual case comes from a hydrocyclone in laboratory scale.

## 2. Construction of information granules

Information granules are collections of entities that are arranged due to their similarity, functional adjacency, or indiscernibility relation. The process of forming information granules is referred to as IG. There are many approaches to construction of IG, for example SOM, Fuzzy C-Means (FCM), and RST. The granulation level depends on the requirements of the project. The smaller IGs come from more detailed processing. On the other hand, because of complex innate feature of information in real world and to deal with vagueness, adopting of fuzzy and rough analysis or the combination form of them is necessary. In this study, the main aim is to develop a hierarchical extraction of IGs using three main steps:

1-Random selection of initial crisp granules: this step can be set as "Close World" Assumption .But in many applications, the assumption of complete information is not feasible (CWA), and only cannot be used. In such cases, an Open World Assumption

(OWA), where information not known by an agent is assumed to be unknown, is often accepted (Dohert et al, 2007).

2- Fuzzy granulation of initial granules: sub fuzzy granules inside precise granules and extraction of if-then rules.

3- The close-open iteration: this process is a guideline to balancing of crisp and sub fuzzy granules by some random selection of initial granules or other optimal structures and increment of supporting rules, gradually. This paper employed two main approaches on constructing of IGs: self organizing feature map as initial granulation, and NFIS as secondary granulation. We called briefly our system SONFIS.

### 2.1. Self Organizing Map-neural network (SOM)

Kohonen self-organizing networks (Kohonen feature maps or topology-preserving maps) are competition-based network paradigm for data clustering. The learning procedure of Kohonen feature maps is similar to the competitive learning networks. The main idea behind competitive learning is simple; the winner takes all. The competitive transfer function returns neural outputs of 0 for all neurons except for the winner which receives the highest net input with output 1.

SOM changes all weight vectors of neurons in the near vicinity of the winner neuron towards the input vector. Due to this property SOM, are used to reduce the dimensionality of complex data (data clustering). Competitive layers will automatically learn to classify input vectors, the classes that the competitive layer finds are depend only on the distances between input vectors (Kohonen, 1990).

### 2.2. Neuro-Fuzzy Inference System (NFIS)

There are different solutions of fuzzy inference systems. Two well-known fuzzy modeling methods are the Tsukamoto fuzzy model and Takagi–Sugeno–Kang (TSK) model. In the present work, only the TSK model has been considered. A typical fuzzy rule in this model is as the following form (equation 1):

$$\text{if } x_i \text{ is } A_1 \text{ and } x_2 \text{ is } A_2 \text{ and } x_n \text{ is } A_n \text{ then } y = f(x) \quad (1)$$

where $f(x)$ is crisp function in the consequent. The function $y=f(x)$ is a polynomial in the input variables $x_1, x_2, ..., x_n$. We will apply here the linear form of this function. For $M$ fuzzy rules of the equation 1, we have $M$ such membership functions $\mu_1, \mu_2, ..., \mu_M$. We assume that each antecedent is followed by the consequent of the linear form as the equation 2:

$$= P_{i0} + \sum_{j=1}^{n} P_{ij} x_j \quad i=1,2,...,M \text{ and } j=1,2,...,n \quad (2)$$

The algebraic product aggregation of the input variables, at the existence of $M$ rules, the Neuro–fuzzy TSK system output signal $y(x)$ upon excitation by the vector $x$ is described by the equation 1.

The adjusted parameters of the system are the nonlinear parameters ($c_j^{(k)}, \sigma_j^{(k)}, b_j^{(k)}$) *for j = 1, 2,..., n and k = 1, 2, ..., M* of the fuzzifier functions and the linear parameters (weights $P_{kj}$) of TSK functions. In contrary to the Mamdani fuzzy inference system, the TSK model generates a crisp output value instead of a fuzzy one. The defuzzifier is not necessary.

The TSK fuzzy inference systems described by equation 3 can be easily implanted in the form of a so called Neuro-fuzzy network structure.

$$y(x) = \frac{1}{\sum_{r=1}^{M}\left[\prod_{j=1}^{n}\mu_r(x_j)\right]} \times \sum_{k=1}^{M}\left(\left[\prod_{J=1}^{n}\mu_k(x_j)\right]\left(p_{k0} + \sum_{j=1}^{n}p_{kj}x_j\right)\right) \quad (3)$$

Figure 1 presents the *5-layer* structure of a Neuro-fuzzy network, realizing the TSK model of the fuzzy system. It is assumed that the functions $y_i$, $y_i = f_i(x)$ are linear of the form (as equation 4)

$$f_i(x) = p_{i0} + \sum_{j=1}^{n} p_{ij} x_j \quad (4)$$

The adaptable parameters of the networks are the variables of the membership functions ($c_j^{(k)}, \sigma_j^{(k)}, b_j^{(k)}$) *for j = 1, 2,..., n and k = 1, 2, ..., M* and the coefficients (linear weights) $p_{ij}$ *for i =1,2,...,M and j =0,1,2,...,n* of the linear Takagi–Sugeno functions. The network in figure 1 has a multilayer form, in which any inputs( x,y), as condition attributes, has two MFs. The details of the procedure can be found in(Jang et al, 1997).

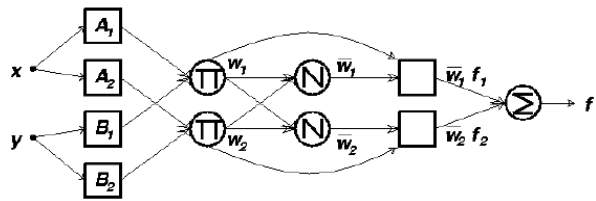

**Figure 1. A typical ANFIS (TSK) with two inputs and two MF for any input (Jang et al, 1997)**

## 3. The proposed methodology

In our algorithm, we use four basic axioms upon the balancing of the successive granules assumption:
Step (1): dividing the monitored data into groups of training and testing data
Step (2): first granulation (crisp) by SOM or other crisp granulation methods
   Step (2-1): selecting the level of granularity randomly or depend on the obtained error from the NFIS or RST (regular neuron growth)
   Step (2-2): construction of the granules (crisp).
Step (3): second granulation (fuzzy or rough IGs) by NFIS or RST
   Step (3-1): crisp granules as a new data.
   Step (3-2): selecting the level of granularity; (Error level, number of rules, strength threshold...)
   Step (3-3): checking the suitability. (Close-open iteration: referring to the real data and reinspect closed world)
   Step (3-4): construction of fuzzy/rough granules.
Step (4): extraction of knowledge rules
Balancing assumption is satisfied by the close-open iterations: this process is a guideline to balancing of crisp and sub fuzzy/rough granules by some random/regular selection of initial granules or other optimal structures and increment of supporting rules (fuzzy partitions or increasing of lower /upper approximations ), gradually.
The overall schematic of Self Organizing Neuro-Fuzzy Inference System -Random and Regular neuron growth-: SONFIS-R, SONFIS-AR; has been shown in figure2.
In first regular granulation, we use a linear relation is given by:
$N_{t+1} = \alpha N_t + \Delta_t; \Delta_t = \beta E_t + \gamma$     (5)

Where $N_t = n_1 \times n_2; |n_1 - n_2| = Min.$ is number of neurons in SOM; $E_t$ is the obtained error (measured error) from second granulation on the test data and coefficients must be

determined, depend on the used data set. Obviously, one can employ like manipulation in the rule (second granulation) generation part, i.e., number of rules.

Determination of granulation level is controlled with three main parameters: range of neuron growth, number of rules and error level. The main benefit of this algorithm is to looking for best structure and rules for two known intelligent system, while in independent situations each of them has some appropriate problems such: finding of spurious patterns for the large data sets, extra-time training of NFIS or SOM.

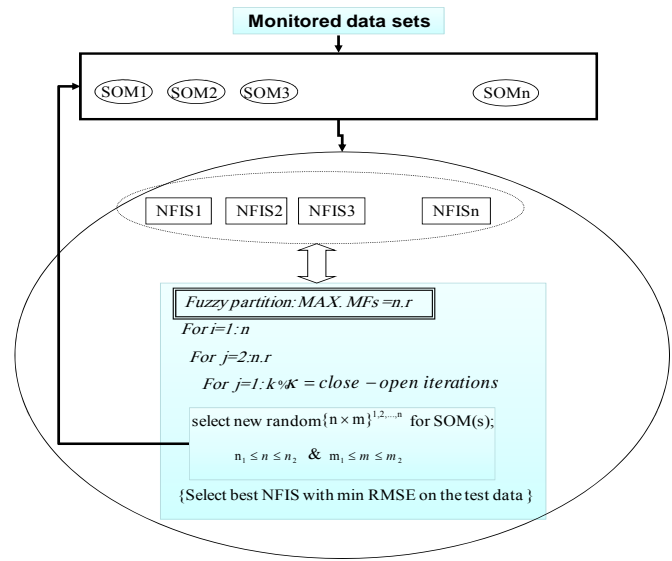

Figure.2. Self Organizing Neuro-Fuzzy Inference System (SONFIS)

Next section, investigates the validation of the proposed algorithm on the monitored available hydrocyclone data, in laboratory scale.

## 4. An example on the hydrocyclone performance

This part of paper investigates applicability and validation of the highlighted methods. For this reason, the following laboratory test on a hydrocyclone has been achieved.

### 4.1. Hydrocyclone

This is a continuously operating classifying device that utilizes centrifugal force to accelerate the settling rate of particles. Its main use in mineral processing is as a classifier, which has proved extremely efficient at fine separation sizes.
 The feed enters tangentially into the cylindrical section of the hydrocyclone and follows a circulating path with a net inward flow of fluid from the outside to the vortex finder on the axis. The high circulating velocities generate large centrifugal fields inside the hydrocyclone. The centrifugal felids usually high enough to create an air core on the axis that usually extends from the spigot (apex) opening at the bottom of the conical section through the vortex finder to the overflow at the top. In order from this to occur the centrifugal force field must be many times larger than the gravitational field. Coarse or high-density particles move rapidly through the fluid tithe outside of the hydrocyclone where they are caught in a downward flow and are removed through the underflow port at the bottom of the hydrocyclone. Fine or low-density particles move more slowly and do not reach the outside of the hydrocyclone. These fine particles are caught in an upward flowing vortex that enters the vortex finder and exit through the overflow port.

There have been numerous efforts to model hydrocyclone. Plitt developed a statistical model to predict the split size, $d_{50}$, of the hydrocyclone .the split size is that particle size that has a 50% chance of exiting in either the overflow or the underflow. Plitt's model has proven to be quite accurate and is often cited in the literature other effective models have been developed but none has been universally adopted, because most of the models are applicable to a limited range of hydrocyclone designs ( Wills,1985 ).

## 4.2. Experimental

Experimental were conducted with the hydrocyclone Test Rig C705 (figure3a).The verification data is divided into hydrocyclone operations according to the different pressure drop (psi) and solid percent, as the tests run with constant geometrical parameters (diameters of the hydrocyclone=50.8 cm, overflow=30m, underflow=7mm). The sample that used in this study were collected from Qara Ağac kaolin mining of Iran, where the special weight of our sample is 2.17 gr/cm$^3$.

The process has four manipulating variables: Pressure drop (psi), solid percent (%), size fraction (μm) and overflow or underflow state (in 0,1codes). The main output of this model is cumulate passing percent (%) that used to control of the split size ($d_{50}$) and, as a direct result, calculation of Imperfection coefficient in the hydrocyclone operations, can be evaluated (figure3b).

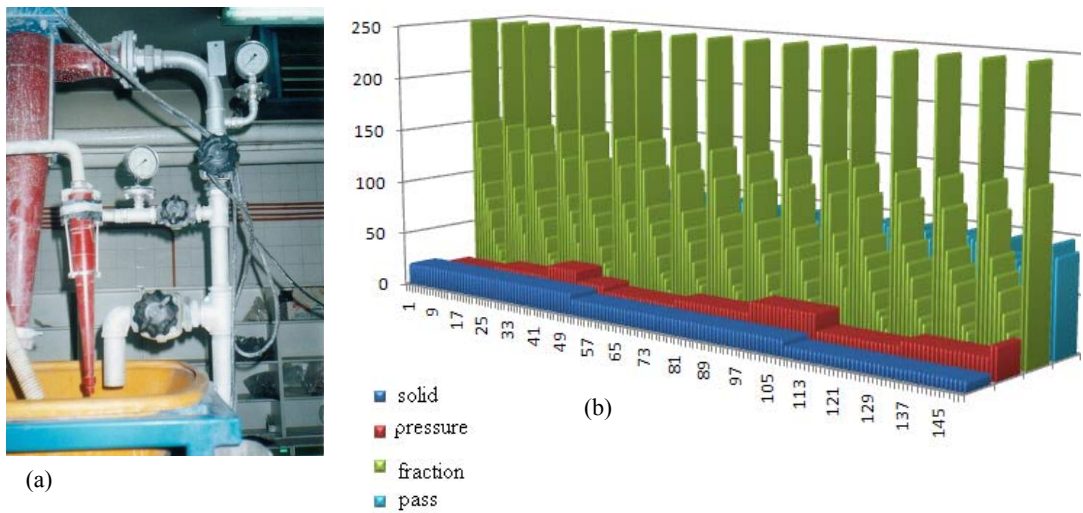

**Figure3.a)The hydrocyclone that used in this study. b) The overall results of the test on the sample**

### 4.3. Results

Analysis of first situation is started off by setting number of close-open iteration and maximum number of rules equal to 10 and 4 in SONFIS-R, respectively. The error measure criteria in SONFIS are Root Mean Square Error (RMSE), given as below:

$$RMSE = \sqrt{\frac{\sum_{i=1}^{m}(t_i - t_i^*)^2}{m}} \; ;$$

Where $t_i$ is output of SONFIS and $t_i^*$ is real answer; m is the number of test data(test objects). In the rest of paper, let m=19 and number of training data set =150. Figures 4 indicates the results of the aforesaid system. The indicated position in figure 4a,b states minimum RMSE over the iterations . figure 5 shows the best crisp granules among 10 itearions for each rule. It is worth noting that upon this balancing criteria ,we may loose the general dominant distribution on the data space. The performance of the obtined fyzzy rules on the test data has been portrayed in figure 6(a). So, the membership functions of each inputs can be compared by thr real training distribution(figure 6b).

By employing of (5) in SONFIS-AR, and $\alpha=1.01$ ; $\beta=.0001$ and $\gamma=.5$ (n.r=2); the general pattern of RMSE vs. neuron growth (in first layer of algorithm) can be observed (figure8a). So,under $\alpha=1.001$ ; $\beta=.001$ and $\gamma=.5$ (n.r=3) the same trends on our system is emerged. It is worth noting that by $\alpha=.9$ , $\beta=.0001$ , $\gamma=.5$ and n.r=2 SONFIS-AR reveals a general chaos form (figure 7). The main reason of this can be followed in the first layer property: regulation of neurons in SOM may get in to the "dead station" and because of random selection of weights in such layer. Other reason is about the range of error vacillation. In fact in this case our system has a low sensivity to the error ( solid SONFIS-AR), and then to the neuron growth.

In this case, we can determine two new balance measures: durability of neurons and distribution of points in a neuron-error space. First measure gets lesser than 20 neurons while in second measure system after 50 iterations falls in the "balance hole" with nearly 65 neurons size.

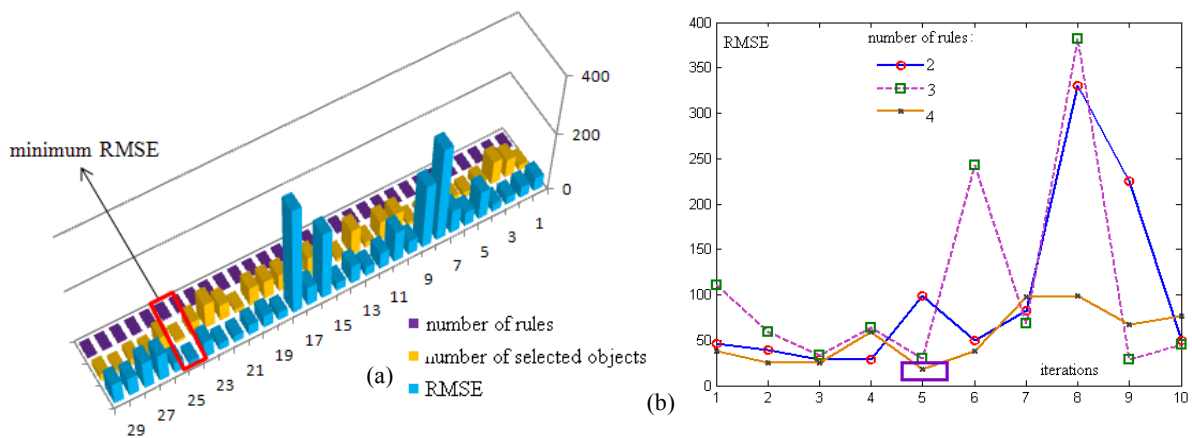

**Figure4.a) obtanined results by SONFIS-R and the minimum RMSE in 30 iteration (10 for each rule). b) the successive RMSE for any rule (2,3 &4)**

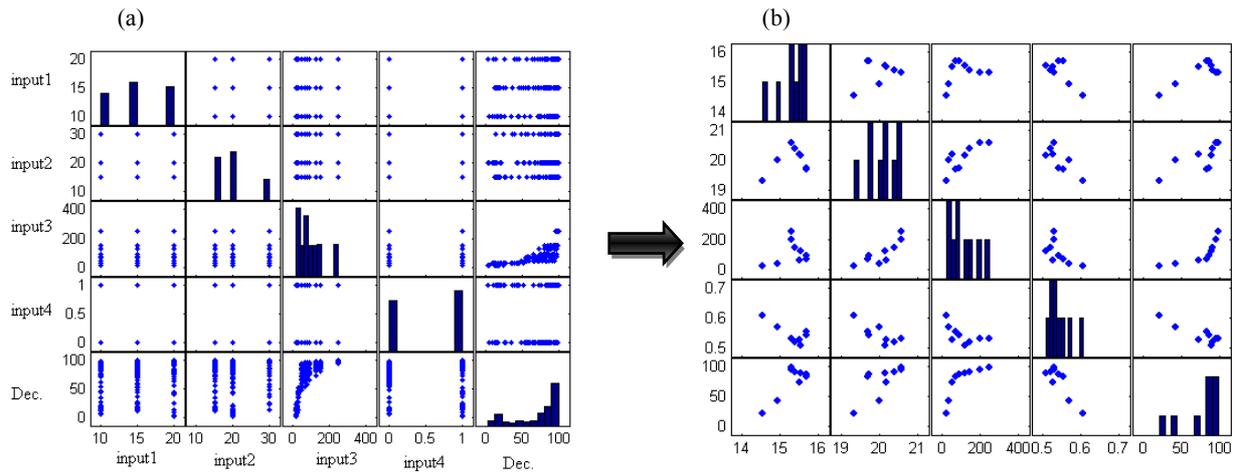

**Figure5.a) Real matrix pot of attributes(training part). b) Reduced crisp atributes using 9*1 SOM- as the best reduced structure by SONFIS-R**

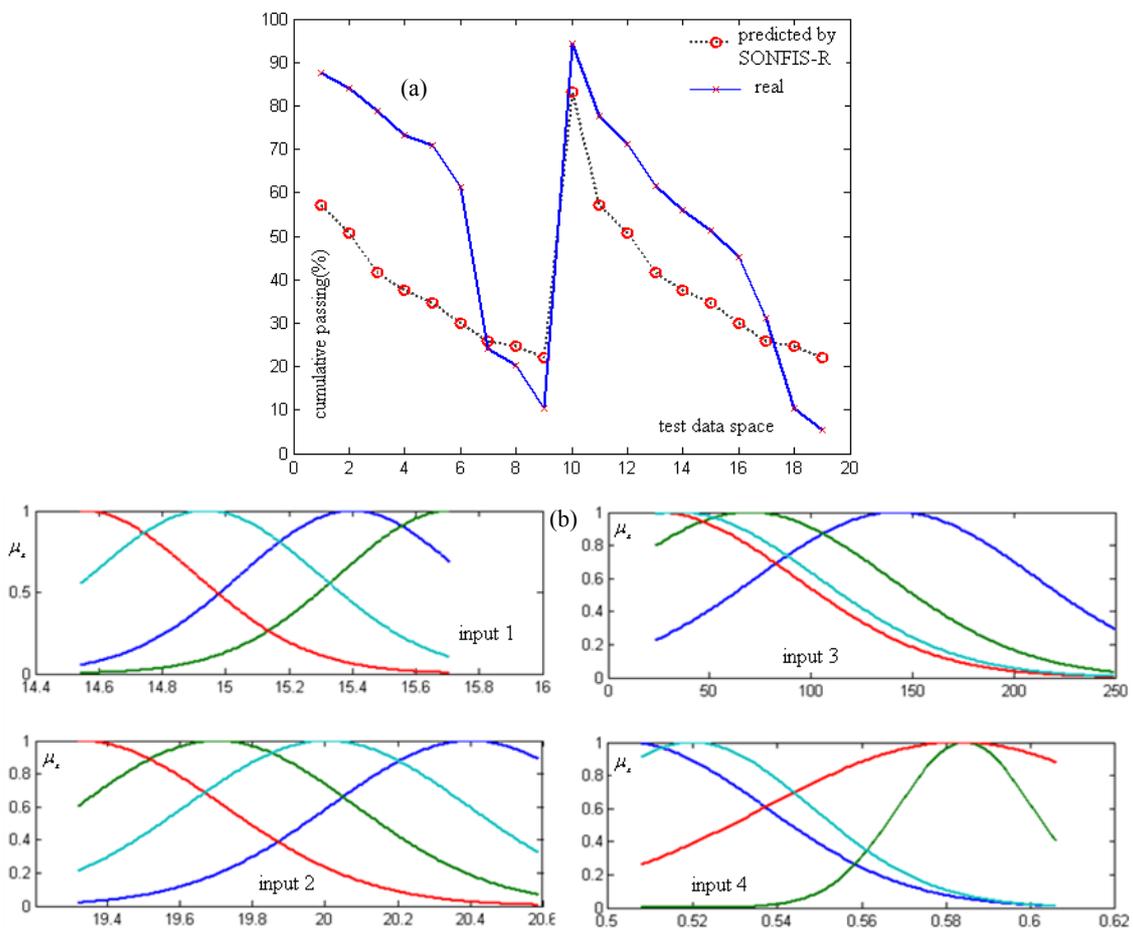

**Figure6.a)the real and predicted decision on the testing data set with sub-fuzzy granulation; b) fuzzy granulation of inputs ;vertical axises are memebership degree( $\mu_x$ )of any input.**

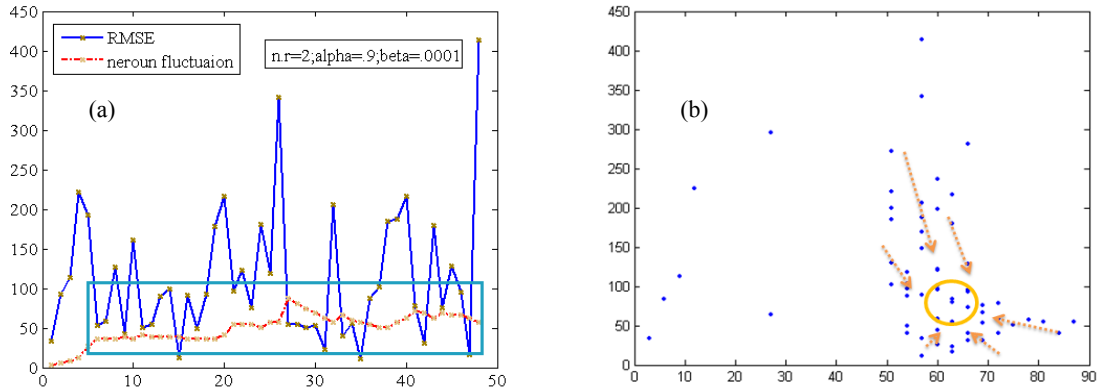

**Figure7 .SONFIS-AR: neuron growth & error fluctuations vs. iteration; $\alpha = .9$ - number of rules =2-a) RMSE-iteration & neuron growth-iteration ; b) RMSE- neuron fluctuation: congestion of points can be used as a "balance trap"**

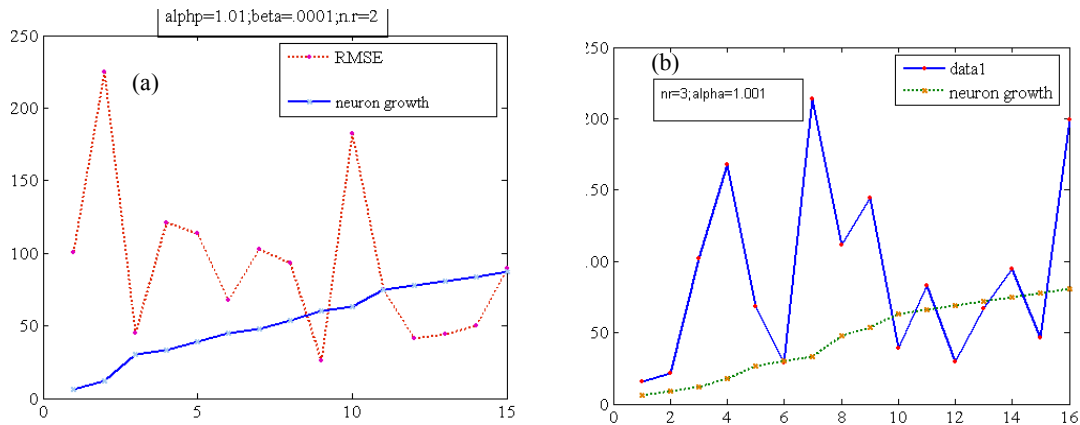

**Figure8. SONFIS-AR: neuron growth & error fluctuations vs. iteration; a) number of rules (*n.r*) =2 , $\alpha = 1.01$ ; b) *n.r*=3 $\alpha = 1.001, \beta = .001$**

## 5. Conclusion

This paper has presented a re-granulation method for knowledge discovery, with emphasis on the crisp-sub fuzzy granules extraction.
This approach has been applied to aid prediction of hydrocyclone performance and extraction of simple rules, which are agents to control of the split size.
The main idea, behind the proposed methodologies, is based on the complexity of information and construction of humanity world cognition by using, as most as, simple rules (in overall structure and number).The competitive between close and open worlds, in a parallel road of the mentioned features, is the additional situation to balancing of the granules and to gain a stable answer.  A collected of granules space, associated with the algorithm's variables, can be supposed as an off-line training of the genetic algorithm, which may find best structure and parameters , in complementary vein on the employed method.  The results proved the applicability of this method in reducing of data set and elicitation of best simple rules, so that, had a relative good answer on the test data.  Authors are developing such algorithms under rough set theory, heuristic optimization methods, balancing and collaborative approaches (owladeghaffari&Babaei,2008) .